# A Bayesian Sampling Approach to Exploration in Reinforcement Learning

**John Asmuth**[†] **Lihong Li**[†] **Michael L. Littman**[†] **Ali Nouri**[†] **David Wingate**[‡]
[†]Department of Computer Science
Rutgers University
Piscataway, NJ 08854
[‡]Computational Cognitive Science Group
Massachusetts Institute of Technology
Cambridge, MA 02143

## Abstract

We present a modular approach to reinforcement learning that uses a Bayesian representation of the uncertainty over models. The approach, **BOSS** (Best of Sampled Set), drives exploration by sampling multiple models from the posterior and selecting actions optimistically. It extends previous work by providing a rule for deciding when to resample and how to combine the models. We show that our algorithm achieves near-optimal reward with high probability with a sample complexity that is low relative to the speed at which the posterior distribution converges during learning. We demonstrate that **BOSS** performs quite favorably compared to state-of-the-art reinforcement-learning approaches and illustrate its flexibility by pairing it with a non-parametric model that generalizes across states.

## 1 INTRODUCTION

The exploration-exploitation dilemma is a defining problem in the field of reinforcement learning (RL). To behave in a way that attains high reward, an agent must acquire experience that reveals the structure of its environment, reducing its uncertainty about the dynamics. A broad spectrum of exploration approaches has been studied, which can be coarsely classified as *belief-lookahead*, *myopic*, and *undirected* approaches.

Belief-lookahead approaches are desirable because they make optimal decisions in the face of their uncertainty. However, they are generally intractable forcing algorithm designers to create approximations that sacrifice optimality. A state-of-the-art belief-lookahead approach is **BEETLE** (Poupart et al., 2006), which plans in the continuous belief space defined by the agent's uncertainty.

Myopic (Wang et al., 2005) approaches make decisions to reduce uncertainty, but they do not explicitly consider how this reduced uncertainty will impact future reward. While myopic approaches can lay no claim to optimality in general, some include guarantees on their total regret or on the number of subtoptimal decisions made during learning. An example of such an algorithm is **RMAX** (Brafman & Tennenholtz, 2002), which distinguishes "known" and "unknown" states based on how often they have been visited. It explores by acting to maximize reward under the assumption that unknown states deliver maximum reward.

Undirected (Thrun, 1992) approaches take exploratory actions, but without regard to what parts of their environment models remain uncertain. Classic approaches such as $\epsilon$-greedy and Boltzmann exploration that choose random actions occasionally fall into this category. The guarantees possible for this class of algorithms are generally weaker—convergence to optimal behavior in the limit, for example. A sophisticated approach that falls into this category is **Bayesian DP** (Strens, 2000). It maintains a Bayesian posterior over models and periodically draws a sample from this distribution. It then acts optimally with respect to this sampled model.

The algorithm proposed in this paper (Section 2) is a myopic Bayesian approach that maintains its uncertainty in the form of a posterior over models. As new information becomes available, it draws a set of samples from this posterior and acts optimistically with respect to this collection—the best of sampled set (or **BOSS**). We show that, with high probability, it takes near-optimal actions on all but a small number of trials (Section 3). We have found that its behavior is quite promising, exploring better than undirected approaches and scaling better than belief-lookahead approaches (Section 4). We also demonstrate its compatibility with sophisticated Bayesian models, resulting in an approach that can generalize experience between states (Section 5).

Note that our analysis assumes a black box algorithm that can sample from a posterior in the appropriate model class. Although great strides have been made recently in representing and sampling from Bayesian posteriors, it remains a challenging and often intractable problem. The correctness of our algorithm also requires that the prior it uses is an accurate description of the space of models—as if the environment is chosen from the algorithm's prior. Some assumption of this form is necessary for a Bayesian approach to show any benefits over an algorithm that makes a worst-case assumption.

## 2 BOSS: BEST OF SAMPLED SET

The idea of sampling from the posterior for decision making has been around for decades (Thompson, 1933). Several recent algorithms have used this technique for Bayesian RL (Strens, 2000; Wilson et al., 2007). In this context, Bayesian posteriors are maintained over the space of Markov decision processes (MDPs) and sampling the posterior requires drawing a complete MDP from this distribution.

Any sampling approach must address a few key questions: 1) When to sample, 2) How many models to sample, and 3) How to combine models. A natural approach to the first question is to resample after every $T$ timesteps, for some fixed $T$. There are challenges to selecting the right value of $T$, however. Small $T$ can lead to "thrashing" behavior in which the agent rapidly switches exploration plans and ends up making little progress. Large $T$ can lead to slow learning as new information in the posterior is not exploited between samples. Strens (2000) advocates a $T$ approximating the depth of exploratory planning required. He suggests several ways to address the third question, leaving their investigation for future work.

**BOSS** provides a novel answer to these questions. It samples multiple models ($K$) from the posterior whenever the number of transitions from a state–action pair reaches a pre-defined threshold ($B$). It then combines the results into an optimistic MDP for decision making—a process we call *merging*. Analogously to **RMAX**, once a state–action pair has been observed $B$ times, we call it *known*.

In what follows, we use $S$ to refer to the size of the state space, $A$ the size of the action space, and $\gamma$ the discount factor. All sampled MDPs share these quantities, but differ in their transition functions. For simplicity, we assume the reward function is known in advance; otherwise, it can be encoded in the transitions.

Given $K$ sampled models from the posterior, $m_1, m_2, \cdots, m_K$, merging is the process of creating a new MDP, $m^\#$, with the same state space, but an augmented action space of $KA$ actions. Each action $a_{i,j}$ in $m^\#$, for $i \in \{1, \cdots, K\}, j \in \{1, \cdots, A\}$, corresponds to the $j$th action in $m_i$. Transition and reward functions are formed straightforwardly—the transition function for $a_{i,j}$ is copied from the one for $a_j$ in $m_i$, for example. Finally, for any state $s$, if a policy in $m^\#$ is to take an action $a_{ij}$, then the actual action taken in the original MDP is $a_j$. A complete description of **BOSS** is given in Algorithm 1.

**Algorithm 1 BOSS** Algorithm

**Inputs:** $K, B$
Initialize the current state $s_1$.
do_sample $\leftarrow$ TRUE.
$q_{s,a} \leftarrow 0, \forall s, a$
**for all** timesteps $t = 1, 2, 3, \ldots$ **do**
  **if** do_sample **then**
    Sample $K$ models $m_1, m_2, \cdots, m_K$ from the posterior (initially, the prior) distribution.
    Merge the models into the mixed MDP $m^\#$.
    Solve $m^\#$ to obtain $\pi_{m^\#}$.
    do_sample $\leftarrow$ FALSE.
  **end if**
  Use $\pi_{m^\#}$ for action selection: $a_t \leftarrow \pi_{m^\#}(s_t)$, and observe reward $r_t$ and next state $s_{t+1}$.
  $q_{s_t,a_t} \leftarrow q_{s_t,a_t} + 1$.
  Update the posterior distribution based the transition $(s_t, a_t, r_t, s_{t+1})$.
  **if** $q_{s_t,a_t} = B$ **then**
    do_sample $\leftarrow$ TRUE
  **end if**
**end for**

**BOSS** solves no more than $SA$ merged MDPs, requiring polynomial time for planning. It draws a maximum of $KSA$ samples. Thus, in distributions in which sampling can be done efficiently, the overall computational demands are relatively low.

## 3 ANALYSIS

This section provides a formal analysis of **BOSS**'s efficiency of exploration. We view the algorithm as a non-stationary policy, for which a value function can be defined. As such, the value of state $s$, when visited by algorithm $\mathcal{A}$ at time $t$, denoted by $V^{\mathcal{A}_t}(s_t)$, is the expected discounted sum of future rewards the algorithm will collect after visiting $s$ at time $t$. Our goal is to show that, when parameters $K$ and $B$ are chosen appropriately, with high probability, $V^{\mathcal{A}_t}(s_t)$ will be $\epsilon$-close to optimal except for a polynomial number of steps (Theorem 3.1). Our objective, and some of our techniques, closely follow work in the PAC-MDP framework (Kakade, 2003; Strehl et al., 2006).

### 3.1 A GENERAL SAMPLE COMPLEXITY BOUND FOR BOSS

Let $m^*$ be the true MDP. When possible, we denote quantities related to this MDP, such as $V^*_{m^*}$, by their shorthand versions, $V^*$. By assumption, the true MDP $m^*$ is drawn from the prior distribution, and so after observing a sequence of transitions, $m^*$ may be viewed as being drawn from the posterior distribution.

**Lemma 3.1** *Let $s_0$ be a fixed state, $p'$ the posterior distribution over MDPs, and $\delta_1 \in (0,1)$. If the sample size $K = \Theta(\frac{1}{\delta_1} \ln \frac{1}{\delta_1})$, then with probability at least $1 - \delta_1$, a model among these $K$ models is optimistic compared to $m^*$ in $s_0$: $\max_i V^*_{m_i}(s_0) \geq V^*(s_0)$.*

**Proof (sketch)**. For any fixed, true model $m^*$, define $P$ as the probability of sampling an optimistic model according to $p'$:

$$P = \sum_{m \in \mathcal{M}} p'(m) \mathbb{I}\left(V^{\pi_m}_m(s_0) \geq V^{\pi_{m^*}}_{m^*}(s_0)\right),$$

where $\mathbb{I}(\cdot)$ is the set-indicator function and $\mathcal{M}$ is the set of MDPs. We consider two mutually exclusive cases. In the first case where $P \geq \delta_1/2$, the probability that none of the $K$ sampled models is optimistic is $(1-P)^K$, which is at most $(1-\delta_1/2)^K$. Let this failure probability $(1-\delta_1/2)^K$ be $\delta_1/2$ and solve for $K$ to get

$$K = \frac{\log(\delta_1/2)}{\log(1-\delta_1/2)} = \Theta\left(\frac{1}{\delta_1}\log\frac{1}{\delta_1}\right).$$

The other case where $P < \delta_1/2$ happens with small probability since the chance of drawing any model, including $m^*$, from that part of the posterior is at most $\delta_1/2$. Combining these two cases, the probability that no optimistic model is included in the $K$ samples is at most $\delta_1/2 + \delta_1/2 = \delta_1$. □

**Lemma 3.2** *The sample size $K = \Theta(\frac{S^2A}{\delta} \ln \frac{SA}{\delta})$ suffices to guarantee $V^*_{m^\#}(s) \geq V^*(s)$ for all $s$ during the entire learning process with probability at least $1-\delta$.*

**Proof (sketch)**. For each model-sampling step, the construction of $m^\#$ implies $V^*_{m^\#}(s) \geq V^*_{m_i}(s)$. By a union bound over all state–action pairs and Lemma 3.1, we have $V^*_{m^\#}(s) \geq V^*(s)$ for all $s$ with probability at least $1 - S\delta_1$. During the entire learning process, there are at most $SA$ model-sampling steps. Applying a union bound again to these steps, we know $V^*_{m^\#}(s) \geq V^*(s)$ for all $s$ in every $K$-sample set with probability at least $1 - S^2A\delta_1$. Letting $\delta = S^2A\delta_1$ completes the proof. □

To simplify analysis, we assume that samples in a state–action pair do not affect the posterior of transition probabilities in other state–actions. However, the result should hold more generally with respect to the posterior induced by the experience in the other states. Define the *Bayesian concentration sample complexity*, $f(s,a,\epsilon,\delta,\rho)$, as the minimum number $c$ such that, if $c$ IID transitions from $(s,a)$ are observed, then with probability $1-\delta$ the following holds true: an $\epsilon$-ball (measured by $\ell_1$-distance) centered at the true model $m^*$ has at least $1-\rho$ probability mass in the posterior distribution. Formally, with probability at least $1-\delta$,

$$\Pr_{m \sim \text{posterior}}\left(\|T_m(s,a) - T_{m^*}(s,a)\|_1 < \epsilon\right) \geq 1-\rho.$$

We call $\rho$ the *diffusion parameter.*

**Lemma 3.3** *If the knownness parameter $B = \max_{s,a} f(s,a,\epsilon,\frac{\delta}{SA},\frac{\rho}{S^2A^2K})$, then the transition function of all the sampled models are $\epsilon$-close (in the $\ell_1$ sense) to the true transition function for all the known state–action pairs during the entire learning process with probability at least $1-\delta-\rho$.*

**Proof (sketch)**. The proof consists of several applications of the union bound. The first is applied to all state–action pairs, implying the posterior concentrates around the true model for all state–action pairs with diffusion $\rho' = \frac{\rho}{S^2A^2K}$ with probability at least $1-\delta$.

Now, suppose the posterior concentrates around $m^*$ with diffusion $\rho'$. For any known $(s,a)$, the probability that a sampled MDP's transition function in $(s,a)$ is $\epsilon$-accurate is at least $1-\rho'$, according to the definition of $f$. By the union bound, the sampled MDP's transition function is $\epsilon$-accurate in all known state–action pairs with probability at least $1 - SA\rho'$. A union bound is applied a second time to the $K$ sampled models, implying all $K$ sampled MDPs' transition functions are $\epsilon$-accurate in all known state–action pairs with probability at least $1 - SAK\rho'$. Finally, using a union bound a third time to all model-sampling steps in **BOSS**, we know that all sampled models have $\epsilon$-accurate transitions in all known $(s,a)$ with probability at least $1 - S^2A^2K\rho' = 1-\rho$. Combining this result with the $\delta$ failure probability in the previous paragraph completes the proof. □

**Theorem 3.1** *When the knownness parameter $B = \max_{s,a} f\left(s,a,\epsilon(1-\gamma)^2,\frac{\delta}{SA},\frac{\delta}{S^2A^2K}\right)$, then with probability at least $1-4\delta$, $V^{\mathcal{A}_t}(s_t) \geq V^*(s_t) - 4\epsilon$ in all but $\zeta(\epsilon,\delta) = O\left(\frac{SAB}{\epsilon(1-\gamma)^2}\ln\frac{1}{\delta}\ln\frac{1}{\epsilon(1-\gamma)}\right)$ steps.*

**Proof (sketch)**. The proof relies on a general PAC-MDP theorem by Strehl et al. (2006) by verifying their three required conditions hold. First, the value function is optimistic, as guaranteed by Lemma 3.2. Second, the accuracy condition is satisfied since the $\ell_1$-error in the transition probabilities, $\epsilon(1-\gamma)^2$, translates into an $\epsilon$ error bound in the value function (Kearns & Singh, 2002). Lastly, the agent visits an unknown state–action at most $SAB$ times, satisfying the learning complexity condition. The probability that any of the three conditions fails is, due to a union bound, at most $3\delta$: the first $\delta$ comes from Lemma 3.2, and the other two from Lemma 3.3. $\square$

### 3.2 THE BAYESIAN CONCENTRATION SAMPLE COMPLEXITY

Theorem 3.1 depends on the Bayesian concentration sample complexity $f$. A full analysis of $f$ is beyond the scope of this paper. In general, $f$ depends on certain properties of the model space as well as the prior distribution. While it is likely that a more accurate estimate of $f$ can be obtained in special cases, we make use of a fairly general result by Zhang (2006) to relate our sample complexity of exploration in Theorem 3.1 to certain characteristics of the Bayesian prior. Future work can instantiate this general result to special MDP classes and prior distributions.

We will need two key quantities introduced by Zhang (2006; Section 5.2). The first is the *critical prior-mass radius*, $\varepsilon_{p,n}$, which characterizes how dense the prior distribution $p$ is around the true model (smaller values imply denser priors). The second is the *critical upper-bracketing radius with coefficient* 2/3, denoted $\varepsilon_{\mathrm{upper},n}$, whose decay rate (as $n$ becomes large) controls the consistency of the Bayesian posterior distribution. When $\varepsilon_{\mathrm{upper},n} = o(1)$, the posterior is consistent. Now, define $\varepsilon_n = 4\varepsilon_{p,n} + \frac{3}{2}\varepsilon_{\mathrm{upper},n}$. The next lemma states that as long as $\varepsilon_n$ decreases sufficiently fast as $n \to \infty$, we may upper bound the Bayesian concentration sample complexity.

**Lemma 3.4** *If there exists a constant* $c > 0$ *such that* $\varepsilon_n = O(n^{-c})$, *then* $f(s, a, \epsilon, \delta, \rho) = \max\{O(\epsilon^{-\frac{2}{c}}\delta^{-\frac{1}{c}}), O(\epsilon^{-2}\delta^{-1}\ln\frac{1}{\rho})\}$.

**Proof (sketch)**. We set $\rho = 1/2$ and $\gamma = 2$ as used in Corollary 5.2 of Zhang (2006) to solve for $n$. Zhang's corollary is stated using Rényi-entropy ($D_{\frac{1}{2}}^{\mathrm{RE}}$) as the distance metric between distributions. But, the same bound applies straightforwardly to $\ell_1$-distance because $D_{\frac{1}{2}}^{\mathrm{RE}}(q||p) \geq \|p - q\|_1^2/2$. $\square$

We may further simplify the result in Lemma 3.4 by assuming without loss of generality that $c \leq 1$, resulting in a potentially looser bound of $f(s, a, \epsilon, \delta, \rho) = O(\epsilon^{-\frac{2}{c}}\delta^{-\frac{1}{c}}\ln\frac{1}{\rho})$. A direct consequence of this simplified result, when combined with Theorem 3.1, is that **BOSS** behaves $\epsilon$-optimally with probability at least $1 - \delta$ in all but at most

$$\tilde{O}\left(\frac{S^{1+\frac{1}{c}}A^{1+\frac{1}{c}}}{\epsilon^{1+\frac{2}{c}}\delta^{\frac{1}{c}}(1-\gamma)^{2+\frac{4}{c}}}\right)$$

steps, where $\tilde{O}(\cdot)$ suppresses logarithmic dependence. This result formalizes the intuition that, if the problem-specific quantity $\varepsilon_n$ decreases sufficiently fast, **BOSS** enjoys polynomial sample complexity of exploration with high probability.

When an uninformative Dirichlet prior is used, it can be shown that $f$ is polynomial in all relevant quantities, and thus Theorem 3.1 provides a performance guarantee similar to the PAC-MDP result for **RMAX** (Kakade, 2003).

## 4 EXPERIMENTS

This section presents computational experiments with **BOSS**, evaluating its performance on a simple domain from the literature to allow for a comparison with other published approaches.

Consider the well-studied 5-state chain problem (Chain) (Strens, 2000; Poupart et al., 2006). The agent has two actions: Action 1 advances the agent along the chain, and Action 2 resets the agent to the first node. Action 1, when taken from the last node, leaves the agent where it is and gives a reward of 10—all other rewards are 0. Action 2 always has a reward of 2. With probability 0.2 the outcomes are switched, however. Optimal behavior is to always choose Action 1 to reach the high reward at the end of the chain.

The slip probability 0.2 is the same for all state–action pairs. Poupart et al. (2006) consider the impact of encoding this constraint as a strong prior on the transition dynamics. That is, whereas in the Full prior, the agent assumes each state–action pair corresponds to independent multinomial distributions over next states, under the Tied prior, the agent knows the underlying transition dynamics except for the value of a single slip probability that is shared between all state–action pairs. They also introduce a Semi prior in which the two actions have independent slip probabilities. Posteriors for Full can be maintained using a Dirichlet (the conjugate for the multinomial) and Tied/Semi can be represented with a simple Beta distribution.

In keeping with published results on this problem, Table 1 reports cumulative rewards in the first 1000 steps, averaged over 500 runs. Standard error is on the order of 20 to 50. The optimal policy for this problem scores 3677. The **exploit** algorithm is one that always acts optimally with respect to the average model weighted by the posterior. **RAM-RMAX** (Leffler et al., 2007) is a version of **RMAX** that can exploit the tied parameters of tasks like this one. Results for **BEETLE** and **exploit** are due to Poupart et al. (2006). All runs used a discount factor of $\gamma = 0.95$ and **BOSS** used $B = 10$ and $K = 5$.

Table 1: Cumulative reward in Chain

| | Tied | Semi | Full |
|---|---|---|---|
| **BEETLE** | 3650 | 3648 | 1754 |
| **exploit** | 3642 | 3257 | 3078 |
| **BOSS** | 3657 | 3651 | 3003 |
| **RAM-RMAX** | 3404 | 3383 | 2810 |

All algorithms perform very well in the Tied scenario (although **RAM-RMAX** is a bit slower as it needs to estimate the slip probability very accurately to avoid finding a suboptimal policy). Poupart et al. (2006) point out that **BEETLE** (a belief-lookahead approach) is more effective than **exploit** (an undirected approach) in the Semi scenario, which requires more careful exploration to perform well. In Full, however, **BEETLE** falls behind because the larger parameter space makes it difficult for it to complete its belief-lookahead analysis.

**BOSS**, on the other hand, explores as effectively as **BEETLE** in Semi, but is also effective in Full. A similarly positive result (3158) in Full is obtained by **Bayesian DP** (Strens, 2000).

# 5 BAYESIAN MODELING OF STATE CLUSTERS

The idea of state clusters is implicit in the Tied prior. We say that two states are in the same cluster if their probability distributions over relative outcomes are the same given any action. In Chain, for example, the outcomes are advancing along the chain or resetting to the beginning. Both actions produce the same distribution on these two outcomes independent of state, Action 1 is 0.8/0.2 and Action 2 is 0.2/0.8, so Chain can be viewed as a one-cluster environment.

We introduce a variant of the chain example, the two-cluster Chain2, which includes an additional state cluster. Cluster 1—states 1, 3, and 5—behaves identically to the cluster in Chain. Cluster 2—states 2 and 4—has roughly the reverse distributions (Action 1 0.3/0.7, Action 2 0.7/0.3).

**RAM-RMAX** can take advantage of cluster structure, but only if it is known in advance. In this section, we show how **BOSS** with an appropriate prior can learn an unknown cluster structure and exploit it to speed up learning.

## 5.1 A NON-PARAMETRIC MODEL OF STATE CLUSTERING

We derive a non-parametric cluster model that can simultaneously use observed transition outcomes to discover which parameters to tie and estimate their values. We first assume that the observed outcomes for each state in a cluster $c$ are generated independently, but from a shared multinomial parameter vector $\theta^c$. We then place a Dirichlet prior over each $\theta^c$ and integrate them out. This process has the effect of coupling all of the states in a particular cluster together, implying that we can use all observed outcomes of states in a cluster to improve our estimates of the associated transition probabilities.

The generative model is

$$\begin{aligned} \kappa &\sim \mathrm{CRP}(\alpha) \\ \theta^{\kappa(s)} &\sim \mathrm{Dirichlet}(\eta) \\ o^{s,a} &\sim \mathrm{Multinomial}(\theta^{\kappa(s)}) \end{aligned}$$

where $\kappa$ is a clustering of states ($\kappa(s)$ is the id of $s$'s cluster), $\theta^{\kappa(s)}$ is a multinomial over outcomes associated with each cluster, and $o^{s,a}$ is the observed outcome counts for state $s$ and action $a$. Here, CRP is a Chinese Restaurant Process (Aldous, 1985), a flexible distribution that allows us to infer both the number of clusters and the assignment of states to clusters. The parameters of the model are $\alpha \in \mathbb{R}$, the concentration parameter of the CRP, and $\eta \in \mathbb{N}^N$, a vector of $N$ pseudo-counts parameterizing the Dirichlet.

The posterior distribution over clusters $\kappa$ and multinomial vectors $\theta$ given our observations $o^{s,a}$ (represented as "data" below) is

$$\begin{aligned} p(\kappa, \theta|\text{data}) &\propto p(\text{data}|\theta)p(\theta|\eta)p(\kappa|\alpha) \\ &= \prod_{s,a} p(o^{s,a}|\theta^{\kappa(s)})p(\theta^{\kappa(s)}|\eta)p(\kappa|\alpha) \\ &= \prod_{c\in\kappa}\prod_{s\in c}\prod_{a\in\mathcal{A}} p(o^{s,a}|\theta^c)p(\theta^c|\eta)p(\kappa|\alpha) \end{aligned}$$

where $c$ is the set of all states in a particular cluster. We now integrate out the multinomial parameter vector $\theta^c$ in closed form, resulting in a standard Dirichlet compound multinomial distribution (or multivariate Polya distribution):

$$p(\text{data}|\kappa) = \prod_{c\in\kappa}\int_{\theta^c}\prod_{s\in c}\prod_{a\in\mathcal{A}} p(o^{s,a}|\theta^c)p(\theta^c|\eta) = \quad (1)$$

$$\prod_{c\in\kappa, a\in\mathcal{A}} \frac{\Gamma(\sum_i \eta_i)}{\prod_i \Gamma(\eta_i)} \frac{\prod_s \Gamma(\sum_i o_i^{s,a}+1)}{\prod_{i,s}\Gamma(\ o_i^{s,a}+1)} \frac{\prod_i \Gamma(\sum_s o_i^{s,a}+\eta_i)}{\Gamma(\sum_{i,s}\ o_i^{s,a}+\eta_i)}.$$

Because the $\theta^c$ parameters have been integrated out of the model, the posterior distribution over models is simply a distribution over $\kappa$. We can also sample transition probabilities for each state by examining the posterior predictive distribution of $\theta^c$.

To sample models from the posterior, we sample cluster assignments and transition probabilities in two stages, using repeated sweeps of Gibbs sampling. For each state $s$, we fix the cluster assignments of all other states and sample over the possible assignments of $s$ (including a new cluster):

$$p(\kappa(s)|\kappa_{-s}, \text{data}) \propto p(\text{data}|\kappa)p(\kappa)$$

where $\kappa(s)$ is the cluster assignment of state $s$ and $\kappa_{-s}$ is the cluster assignments of all other states. Here, $p(\text{data}|\kappa)$ is given by Eq. 1 and $p(\kappa)$ is the CRP prior

$$p(\kappa|\alpha) = \alpha^r \frac{\Gamma(\alpha)}{\Gamma(\alpha + \sum_i \kappa_i)} \prod_{i=1}^{r} \Gamma(\kappa_i)$$

with $r$ the total number of clusters and $\kappa_i$ the number of states in each cluster.

Given $\kappa$, we sample transition probabilities for each action from the posterior predictive distribution over $\theta^c$, which, due to conjugacy, is a Dirichlet distribution:

$$\theta^c|\kappa, \eta, \alpha, a \sim \text{Dirichlet}(\eta + \sum_{s \in c} o^{s,a}).$$

### 5.2 BOSS WITH CLUSTERING PRIOR

We ran **BOSS** in a factorial design where we varied the environment (Chain vs. Chain2) and the prior (Tied, Full, vs. Cluster, where Cluster is the model described in the previous subsection). For our experiments, **BOSS** used a discount factor of $\gamma = 0.95$, knownness parameter $B = 10$, and a sample size of $K = 5$. The Cluster CRP used $\alpha = 0.5$ and whenever a sample was required, the Gibbs sampler ran for a burn period of 500 sweeps with 50 sweeps between each sample.

Figure 1 displays the results of running **BOSS** with different priors in Chain and Chain2. The top line on the graph corresponds to the results for Chain. Moving from left to right, **BOSS** is run with weaker priors—Tied, Cluster, and Full. Not surprisingly, performance decreases with weaker priors. Interestingly, however, Cluster is not significantly worse than Tied—it is able to identify the single cluster and learn it quickly.

The second line on the plot is the results for Chain2, which has two clusters. Here, Tied's assumption of the existence of a single cluster is violated and performance suffers as a result. Cluster outperforms Full by a smaller margin, here. Learning two independent clusters is still better than learning all states separately, but the gap is narrowing. On a larger example with more sharing, we'd expect the difference to be more dramatic. Nonetheless, the differences here are statistically significant ($2 \times 3$ ANOVA $p < 0.001$).

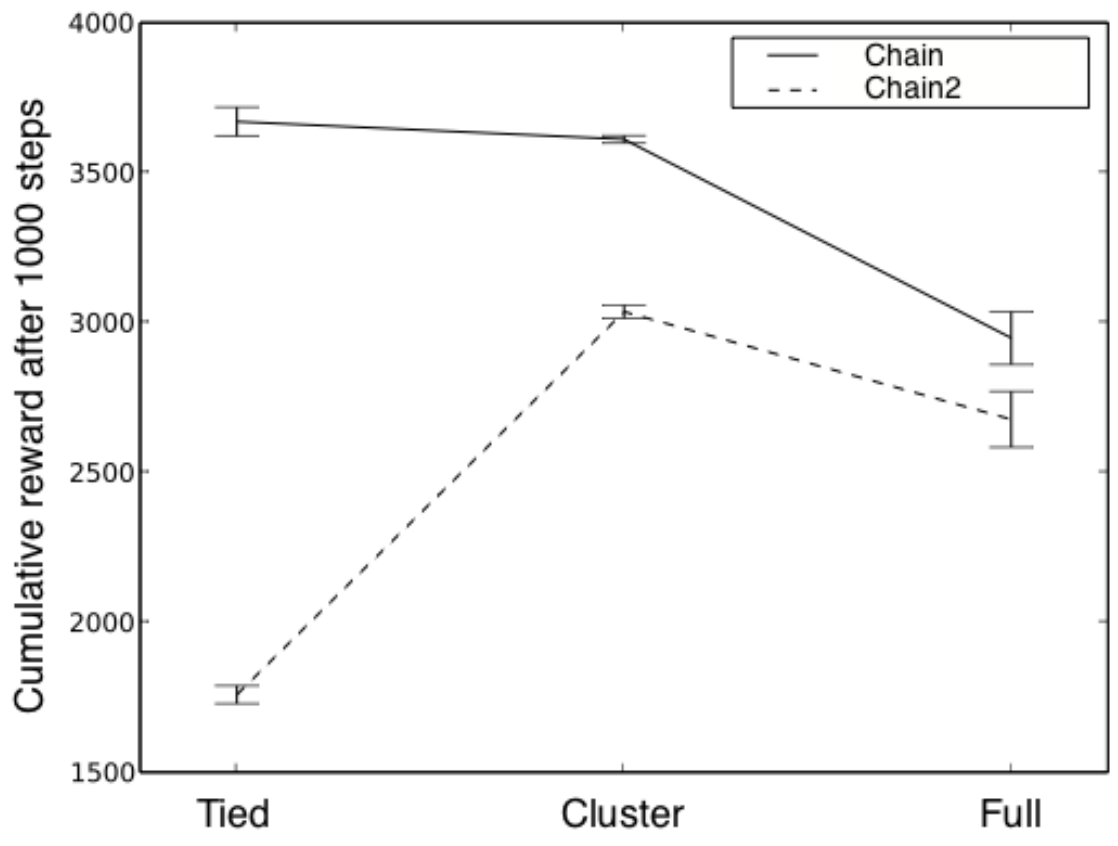

Figure 1: Varying priors and environments in **BOSS**.

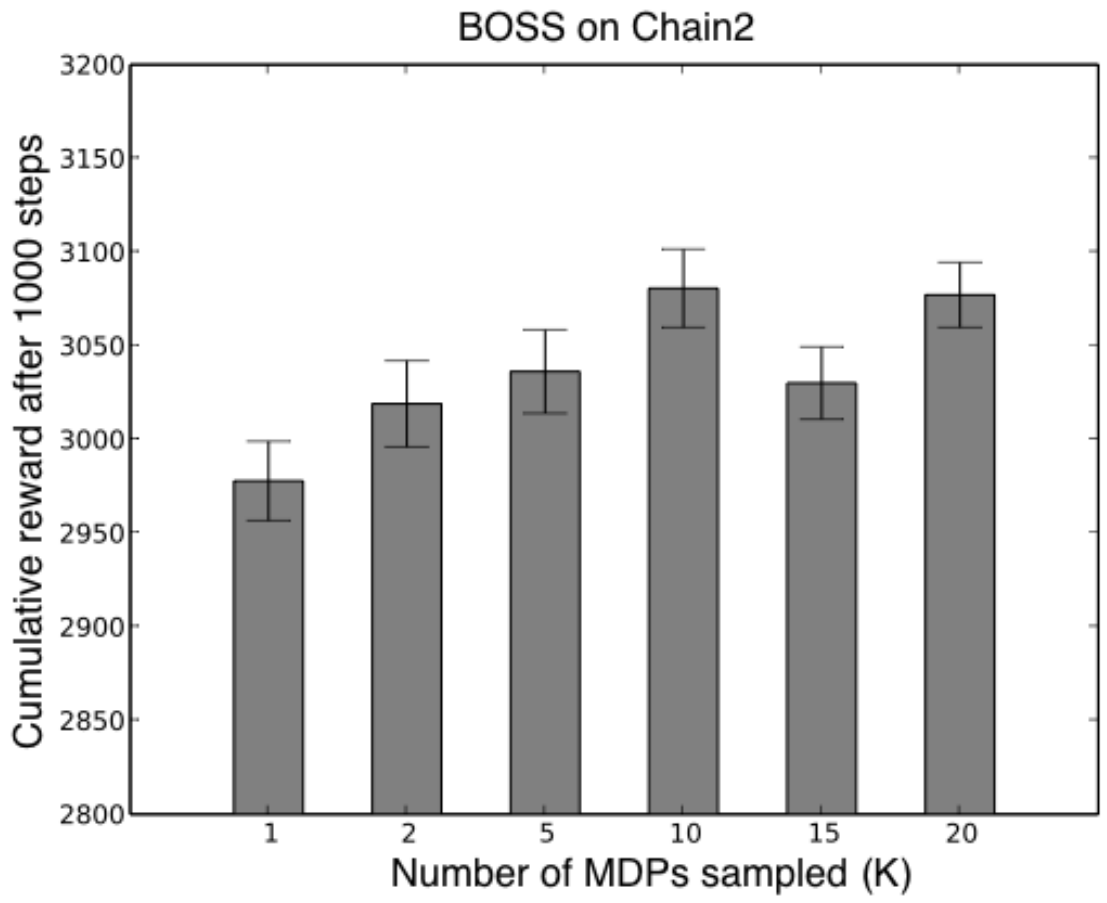

Figure 2: Varying $K$ in **BOSS**.

### 5.3 VARYING $K$

The experiments reported in the previous section used model samples of size $K = 5$. Our next experiment was intended to show the effect of varying the sample size. Note that **Bayesian DP** is very similar to **BOSS** with $K = 1$, so it is important to quantify the impact of this parameter to understand the relationship between these algorithms.

Figure 2 shows the result of running **BOSS** on Chain2 using the same parameters as in the previous section. Note that performance generally improves with $K$. The difference between $K = 1$ and $K = 10$ is statistically significant (t-test $p < 0.001$).

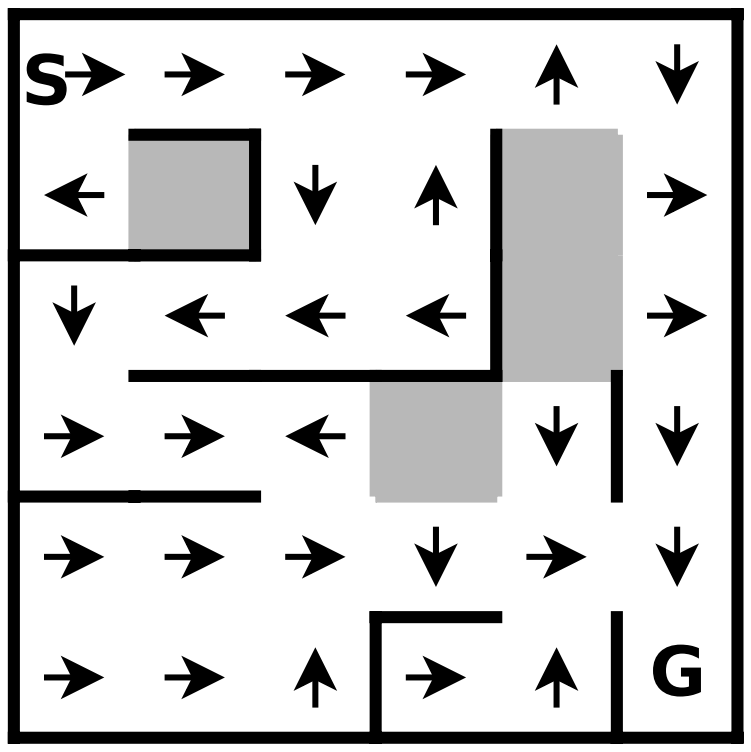

Figure 3: Diagram of 6x6 Marble Maze.

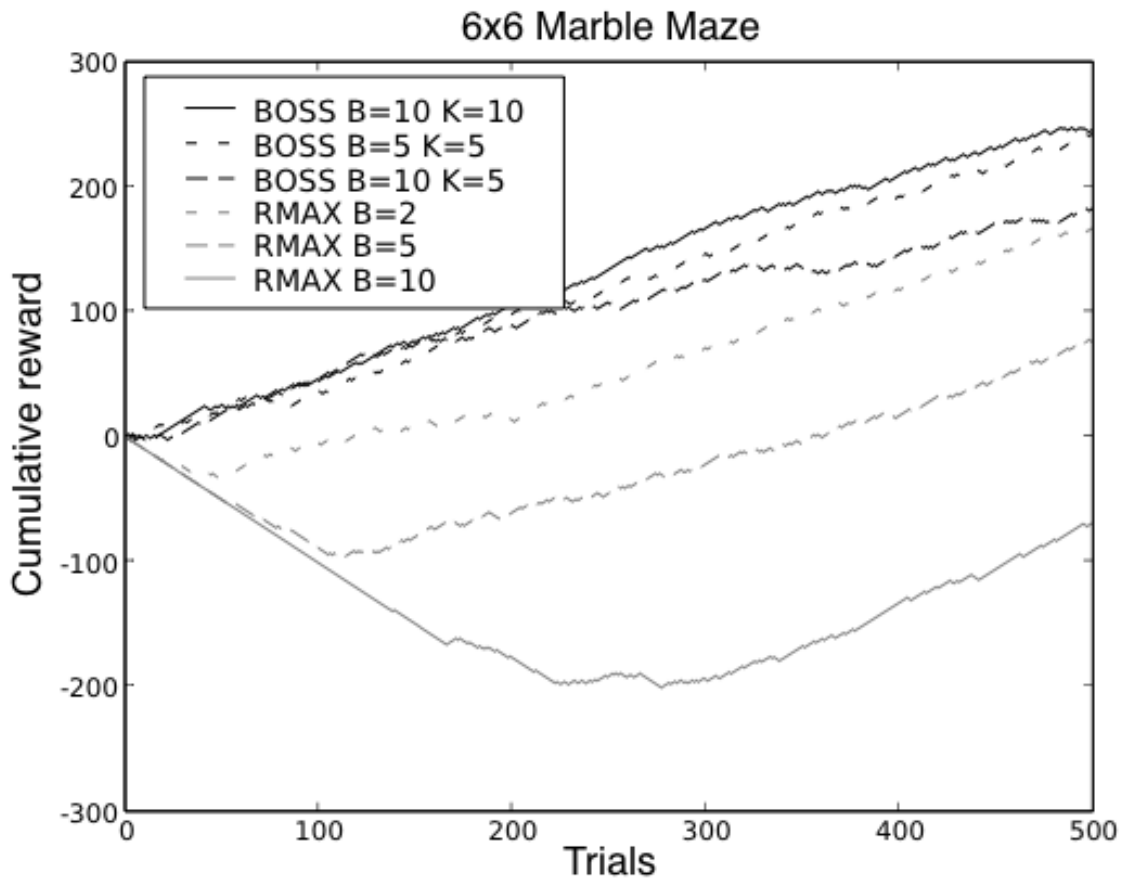

Figure 4: Comparison of algorithms on 6x6 Marble Maze.

### 5.4 6x6 MARBLE MAZE

To demonstrate the exploration behavior of our algorithm, we developed a 6x6 grid-world domain with standard dynamics (Russell & Norvig, 1994). In this environment, the four actions, *N*, *S*, *E* and *W*, carry the agent through the maze on its way to the goal. Each action has its intended effect with probability .8, and the rest of the time the agent travels in one of the two perpendicular directions with equal likelihood. If there is a wall in the direction the agent tried to go, it will remain where it is. Each step has a cost of 0.001, and terminal rewards of $-1$ and $+1$ are received for falling into a pit or reaching the goal, respectively. The map of the domain, along with its optimal policy, is illustrated in Figure 3.

The dynamics of this environment are such that each local pattern of walls (at most 16) can be modeled as a separate cluster. In fact, fewer than 16 clusters appear in the grid and fewer still are likely to be encountered along an optimal trajectory. Nonetheless, we expected **BOSS** to find and use a larger set of clusters than in the previous experiments.

For this domain, **BOSS** used a discount factor of $\gamma = 0.95$ and a CRP hyperparameter of $\alpha = 10$. Whenever an MDP set was needed, the Gibbs sampler ran for a burn period of 100 sweeps with 50 sweeps between each sample. We also ran **RMAX** in this domain.

The cumulative reward achieved by the **BOSS** variants that learned the cluster structure, in Figure 4, dominated those of **RMAX**, which did not know the cluster structure. The primary difference visible in the graph is the time needed to obtain the optimal policy. Remarkably, **BOSS** $B = 10$ $K = 10$ latches onto near optimal behavior nearly instantaneously whereas the **RMAX** variants required 50 to 250 trials before behaving as well. This finding can be partially explained by the choice of the clustering prior and the outcomes it drew from, which effectively put a lower bound on the number of steps to the goal from any state. This information made it easy for the agent to ignore longer paths when it had already found something that worked.

Looking at the clustering performed by the algorithm, a number of interesting features emerge. Although it does not find a one-to-one mapping from states to patterns of walls, it gets very close. In particular, among the states that are visited often in the optimal policy and for the actions chosen in these states, the algorithm groups them perfectly. The first, third, fourth, and fifth states in the top row of the grid are all assigned to the same cluster. These are the states in which there is a wall above and none below or right, impacting the success probability of $N$ and $E$, the two actions chosen in these states. The first, second, third, and fifth states in the rightmost column are similarly grouped together. These are the states with a wall to the right, but none below or left, impacting the success probability of $S$ and $E$, the two actions chosen in these states. Other, less commonly visited states, are clustered somewhat more haphazardly, as it was not necessary to visit them often to obtain high reward in this grid. The sampled models used around 10 clusters to capture the dynamics.

### 5.5 COMPUTATIONAL COMPLEXITY

The computation time required by BOSS depends on two distinct factors. First, the time required for per-step planning using value iteration scales with the number of sampled MDPs, $K$. Second, the time required for sampling new MDPs depends linearly on $K$ and on the type of prior used. For a simple prior, such as Full, samples can be drawn extremely quickly. For a more complex prior, such as Cluster, samples can take longer. In the 6x6 Marble Maze, samples were drawn at a rate of roughly one every ten seconds. It is worth noting that sampling can be carried out in parallel.

# 6 CONCLUSIONS

We presented a modular approach to exploration called **BOSS** that interfaces a Bayesian model learner to an algorithm that samples models and constructs exploring behavior that converges quickly to near optimality. We compared the algorithm to several state-of-the-art exploration approaches and showed it was as good as the best known algorithm in each scenario tested. We also derived a non-parametric Bayesian clustering model and showed how **BOSS** could use it to learn more quickly than could non-generalizing comparison algorithms.

In future work, we plan to analyze the more general setting in which priors are assumed to be only approximate indicators of the real distribution over environments. We are also interested in hierarchical approaches that can learn, in a transfer-like setting, more accurate priors. Highly related work in this direction was presented by Wilson et al. (2007).

An interesting direction for future research is to consider extensions of our clustered state model where the clustering is done in feature space, possibly using non-parametric models such as the Indian Buffet Process (Griffiths & Ghahramani, 2006). Such a model could simultaneously learn how to decompose states into features and also discover which observable features of a state (color, texture, position) are reliable indicators of the dynamics.

We feel that decomposing the details of the Bayesian model from the exploration and decision-making components allow for a very general RL approach. Newly developed languages for specifying Bayesian models (Goodman et al., 2008) could be integrated directly with **BOSS** to produce a flexible learning toolkit.

### Acknowledgements

We thank Josh Tenenbaum, Tong Zhang, and the reviewers. This work was supported by DARPA IPTO FA8750-05-2-0249 and NSF IIS-0713435.